\documentclass{article}
\usepackage{arxiv}
\usepackage[T1]{fontenc}    
\usepackage{hyperref}       
\usepackage{url}            
\usepackage{booktabs}       
\usepackage{amsfonts}       
\usepackage{nicefrac}       
\usepackage{microtype}      
\usepackage{amsmath}
\usepackage{cleveref}       
\usepackage{lipsum}         
\usepackage{graphicx}
\usepackage[authoryear]{natbib}
\usepackage{doi}
\usepackage{hyperref}
\usepackage{graphicx}
\usepackage{latexsym}
\usepackage[ruled, vlined, linesnumbered]{algorithm2e}
\usepackage{tabularx}
\usepackage{booktabs}
\usepackage{authblk} 
\usepackage[semibold]{sourcesanspro} 
\usepackage{multirow}

\title{Generating Effective Ensembles for Sentiment Analysis}

\author[1]{Itay Etelis}
\author[2]{Avi Rosenfeld}
\author[3]{Abraham Itzhak Weinberg}
\author[1]{David Sarne}

\affil[1]{Department of Computer Science, Bar-Ilan University, Israel}
\affil[2]{Department of Computer Science, Jerusalem College of Technology, Israel}
\affil[3]{AI-Weinberg, AI Experts,  Israel}
\begin{document}
\maketitle

\begin{abstract}
In recent years, transformer models have revolutionized Natural Language Processing (NLP), achieving exceptional results across various tasks, including Sentiment Analysis (SA). As such, current state-of-the-art approaches for SA predominantly rely on transformer models alone, achieving impressive accuracy levels on benchmark datasets. In this paper, we show that the key for further improving the accuracy of such ensembles for SA is to include not only transformers, but also traditional NLP models, despite the inferiority of the latter compared to transformer models. However, as we empirically show, this necessitates a change in how the ensemble is constructed, specifically relying on the Hierarchical Ensemble Construction (HEC) algorithm we present. Our empirical studies across eight canonical SA datasets reveal that ensembles incorporating a mix of model types, structured via HEC, significantly outperform traditional ensembles.  Finally, we provide a comparative analysis of the performance of the HEC and GPT-4, demonstrating that while GPT-4 closely approaches state-of-the-art SA methods, it remains outperformed by our proposed ensemble strategy.
\end{abstract}

\noindent

\begin{figure}[ht!]
    \centering
    \begin{minipage}{0.48\textwidth} 
        \centering
        \includegraphics[width=\textwidth]{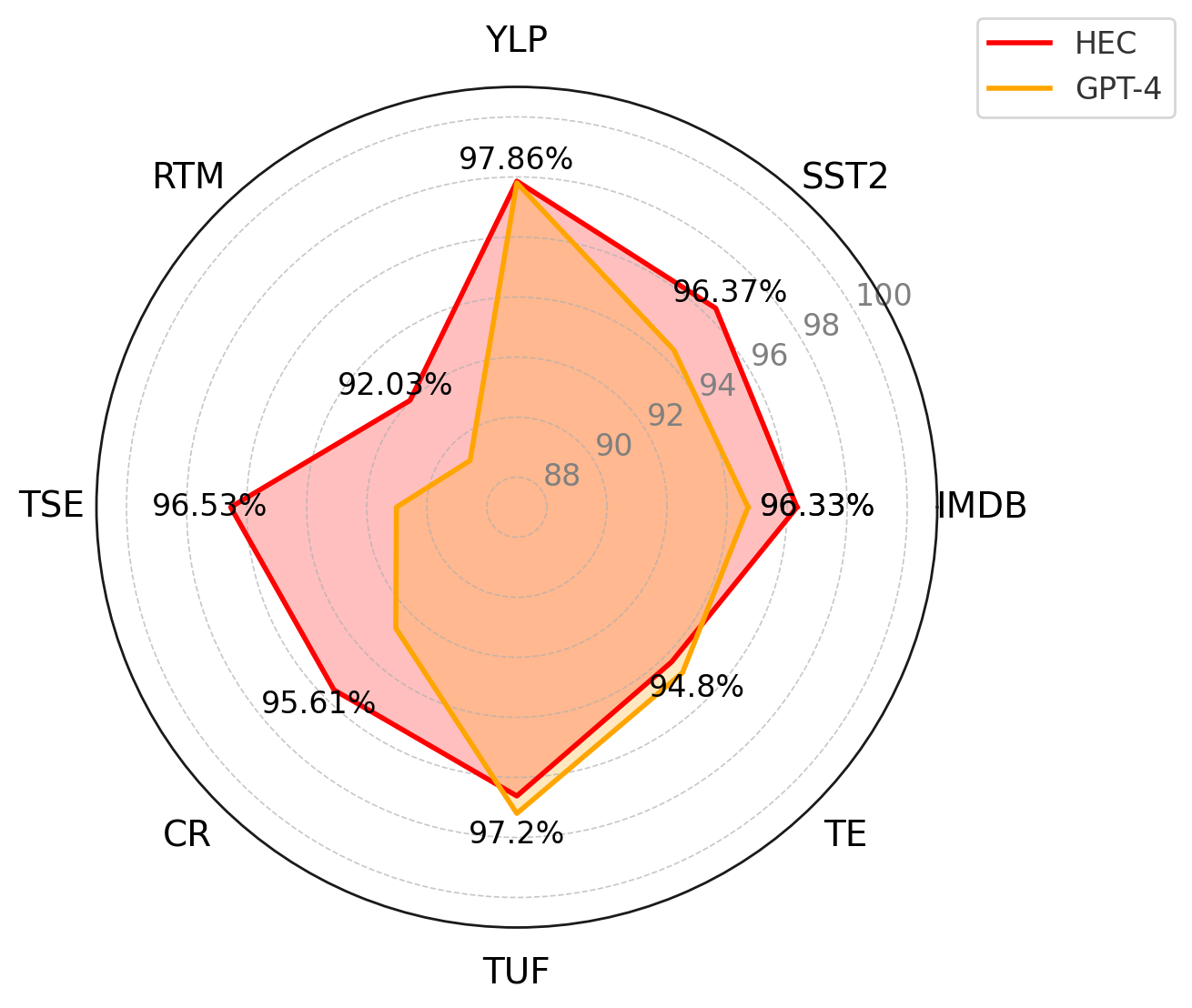}
    \end{minipage}
    \hfill 
    \begin{minipage}{0.48\textwidth} 
        \centering
        \includegraphics[width=\textwidth]{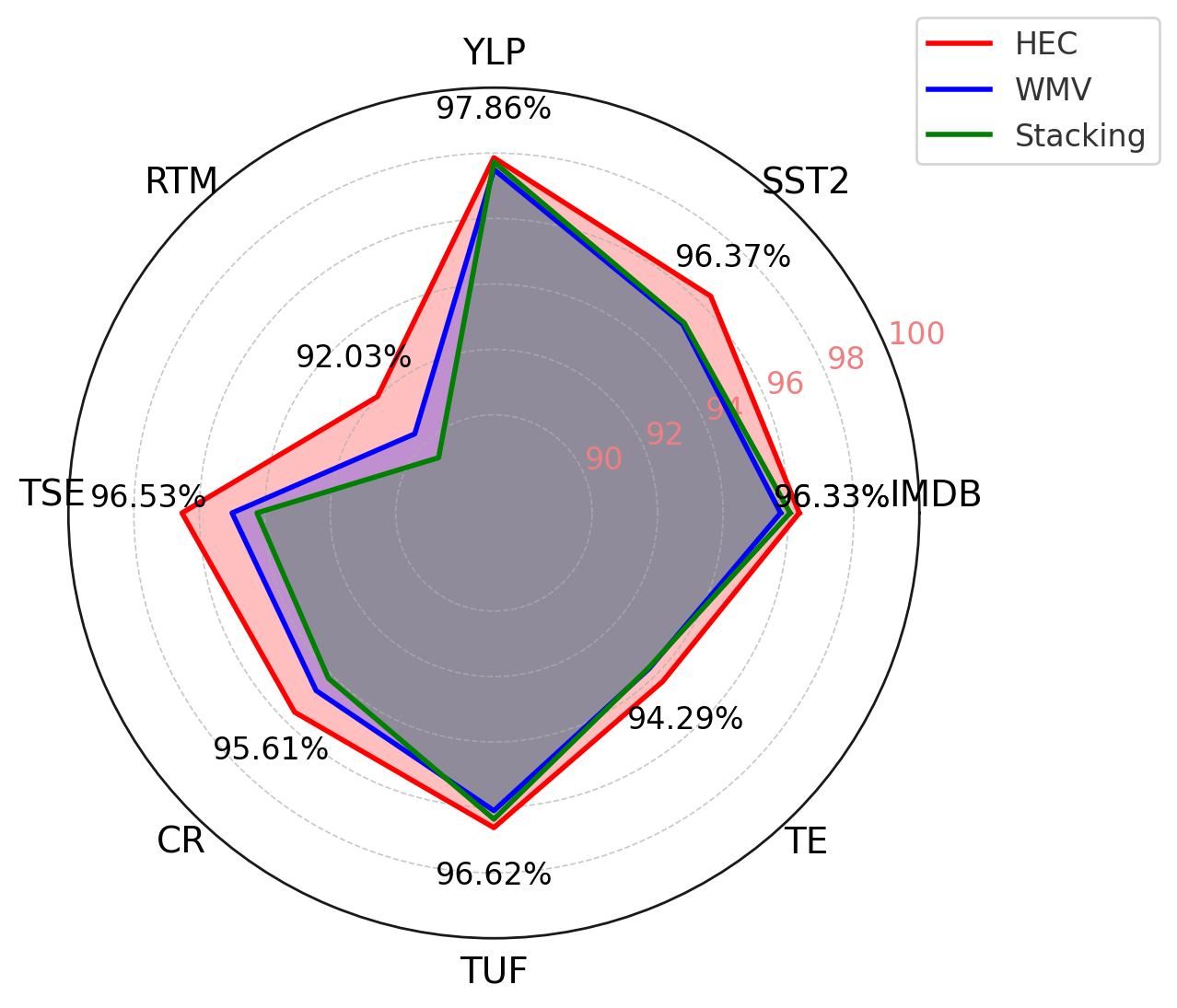}
    \end{minipage}
    \caption{Comparative performance analysis of Hierarchical Ensemble Construction (HEC) against GPT-4 and other ensemble methods (Weighted Majority Voting - WMV, Stacking) over eight datasets. The left chart demonstrates HEC's superior accuracy in six out of eight datasets when compared to GPT-4. The right chart illustrates HEC's consistent improvement across all datasets relative to WMV and Stacking methods. }
    \label{fig:hec_performance}
\end{figure}


\section{Introduction}

Sentiment analysis (SA) is an important NLP task that focuses on identifying and extracting opinions regarding a specific product, topic, or phrase by classifying them into positive, and negative polarities, assigning a score or aggregated ranking, or providing star ratings \citep{Ma2018}. Over the years, SA has gained significant traction as a widely used analytical technique among researchers and various industries, including business, finance, politics, education, and services \citep{liu2010sentiment}. This is due to its capability to classify the emotional content present in unstructured text data as positive or negative and to quantify the intensity or degree of these emotions. As such, SA serves as a beneficial solution for decision-making for policy leaders, business professionals, and service providers \citep{SANCHEZRADA2019344}.  
Many approaches have been suggested for SA classification. They can be broadly classified as lexicon-based, bag-of-word traditional Machine Learning (ML), neural network models with word embedding, and transformer models \citep{hussein2018survey}. Transformer models currently outperform all other options, often by significant amounts \citep{chatgpt_Transformers_vs_lexical}.


Ensemble techniques have been proven to be highly effective in NLP, and also specifically for SA \citep{WANG201477}. Ensembles in ML are systems that combine multiple base-learners (denoted \emph{base-learners} onward) to improve the accuracy and robustness of predictions. This approach resembles the process of consulting various perspectives prior to finalizing a decision, wherein each base-learner offers its unique insight, effectively compensating for the shortcomings of individual base-learners \citep{SAGIOMER2018}.  Applying ensembles to SA is particularly beneficial due to the complexity and variability of human language \citep{KAZMAIER2022115819}. Alas, prior attempts to use ensembles in SA are typically based on incorporating base-learners from a single category out of the four categories of lexicon-based, bag-of-words, word embeddings, and transformer models mentioned above \citep{Parveen2020,ARAQUE2017236,haralabopoulos2020ensemble, ankit, 10379619, danyal2024proposing, Muhammad}.\footnote{See the following section for two exceptions.}  The tendency to use base-learners of the same class in SA ensembles is a consequence of the field's evolutionary progress. As newer methodologies emerged which surpass their predecessors in performance, they naturally became the preferred choice for ensemble construction. In particular, the shift from lexicon-based methods to traditional ML, and subsequently to neural networks with word embeddings, reflected significant leaps in analytical capabilities. Each advancement set a new benchmark for accuracy, leading researchers to favor the most current technology for base-learners. This pattern persisted with the advent of transformer models, where their superior ability to understand language nuances made them the dominant choice for state-of-the-art ensemble methods, ensuring that ensembles remain at the forefront of SA technology.


In this paper, we propose the use of ensembles for SA that do not rely solely on transformers, but rather also make use of base-learners from the other three categories. Indeed the foundational premise of ensembles is based on the belief that optimal results are obtained by capitalizing on the strengths of individual component base-learners while minimizing their conflicting aspects \citep{ARAQUE2017236,da2014tweet}. While transformers are by far the strongest best base-learners, we show that depending only on this homogeneous group of base-learners uses redundant methods and yields diminished effectiveness. This redundancy can sometimes lead to incorrect predictions. The main strength of ensembles that include base-learners from varied categories is therefore in allowing transformers to predominantly guide the analysis in instances where their consensus is strong, while in situations where transformers are prone to error and produce conflicting predictions, the other base-learners play a pivotal role in steering the predictions towards more accurate outcomes. 


Our empirical evaluation across eight canonical SA datasets reveals that ensembles incorporating base-learners from all four categories perform better than similar ensembles which only use transformers in several state-of-the-art ensemble construction methods (specifically WMV, Stacking, Shapley Values, and Bayesian Networks). Furthermore, the performance of these ensembles constructed from transformers show a small improvement compared to an ensemble constructed by randomly picking any transformer. We attribute these rather moderate improvements to the fact that these ensemble construction methods consider all base-learners, giving them different weights or using them as attributes in developing an ML-based prediction model. This may lead to cases where the inclusion of some base-learners together with others that are actually detrimental, and hence should be avoided. 

As an alternative to current state-of-the-art ensemble construction methods for SA, we propose the Hierarchical Ensemble
Construction (HEC) algorithm. HEC differs from conventional ensemble methods by employing a hierarchical approach to select a relatively small subset of base-learners out of the full set available, relying only this subset to develop the prediction model. It begins with an initial small subset of base-learners and iteratively expands this group, aiming to identify the most effective combination of base-learners. The final ensemble is then aggregated as a weighted sum of the different base-learners' votes. We show that the HEC algorithm significantly outperforms all the other methods for SA in the set of eight canonical SA datasets, achieving a mean accuracy of 95.71\%. Furthermore, it surpasses all of the ensembles generated with the other ensemble construction methods, even when enabling them to utilize the full set of base-learners (rather than relying exclusively on transformers). In fact, the Hec ensemble's improvement over the base-learners is 3-4.4 times more than that in other leading ensembles (WMV / Stacking). 

Finally, we compare the performance of HEC to that of GPT-4, which is known for its advanced ability to understand and interpret nuanced language and context, essential for accurately discerning emotional tones in text \citep{wang2023emotional}. We find that HEC consistently surpasses GPT-4 in overall performance, despite GPT-4's advanced language capabilities. 

\section{Related Work}

\label{Ensembles Transformers}
An ensemble is a predictive model that combines the predictions of multiple `weak' or smaller base-learners, to form a `strong' classifier \citep{KAZMAIER2022115819}. The underlying assumption of this approach is that better performance is achieved by leveraging the component base-learners' strengths and lowering conflicts \citep{ARAQUE2017236,da2014tweet}. 
Formally, an ensemble learning model $\phi$ uses an aggregation function $G$ to combine $K$ base-learners ${f_1, f_2,..., f_k}$ to make a single output prediction for each example in the datasets. The prediction for the $i_{th}$ example $x_i$, denoted $y_{i}$, is calculated as follows:
\begin{equation}
    y_{i} = \phi(x_i) = G(f_1(x_i), f_2(x_i),..., f_k(x_i)).
\label{mvsignequation1}
\end{equation}
The specific form of the aggregation function $G$ depends on the ensemble model and the nature of the problem \citep{SAGIOMER2018}.

Selecting the right set of base-learners to be used in the ensemble, and the way their outputs should be aggregated, is complex. Given an initial set of base-learners to choose from, the number of possible subsets that can be used increases exponentially with the size of the set.  Hence, ensemble pruning methods that can exclude some of the base-learners, without jeopardizing much the ensemble performance, can be useful \citep{ONAN2017814}. Much like with pruning methods, our proposed ensemble construction method, HEC, aims for a reduced subset of base-learners to be used in the ensemble. However, while pruning methods start with the full set and gradually remove base-learners, HEC starts from scratch and gradually builds the subset to be used. 

This difference is not merely semantic, but rather carries important implications. With HEC, each base-learner is added based on its ability to improve the ensemble's performance. This step-by-step approach helps in identifying the most complementary base-learner, ensuring that each new addition provides value. Pruning, on the other hand, primarily aims for lowering computational demands (training time and prediction time), simplifying model maintenance and enhancing interpretability. The removal of a base-learner in a pruning method, therefore, is not for improving the ensemble performance but rather primarily for decreasing the set size \citep{GUO2018237}. 

Furthermore, starting with a full ensemble and then pruning can be less efficient due to the `curse of dimensionality'.  When there are too many base-learners, it becomes difficult to accurately assess the contribution of each base-learner due to the increased complexity and potential for overfitting \citep{rosenfeld2012combining}. A smaller set of base-learners selected from scratch avoids this issue.

Two categories of output aggregation methods exist: weighting and meta-learning approaches. Weighting approaches assign different levels of importance to the predictions made by the individual base-learner in an ensemble \citep{SAGIOMER2018}. This approach is often used when the base models have comparable performance. One simple weighting method, known as majority voting, involves choosing the class with the most votes as the final prediction in classification problems \citep{a13040083, HEIKAL2018114}. Another, yet more diverse, method involves assigning weights to the base-learners based on their performance on a validation set. Higher weights are given to base-learners with better performance and lower to base-learners with weaker performance \citep{10.1007/978-3-642-41184-7_23, saeed2022}. 
For example, Dogan and Birant \citep{DOGAN2021114060} suggest setting an initial weight based on the accuracy of each learner in the ensemble when applied on a validation set, and then adjusting the weights using a reward mechanism, such that base-learners that accurately classify observations miss-classified by most other base-learners in the ensemble are given greater weight.
The common alternative to weighting output aggregation suggests combining the predictions of multiple base-learners using a meta-classifier \citep{gaye2021tweet}. 
For example, the \emph{stacking} method divides the dataset into $K$ equal parts (folds) \citep{gaye2021tweet} applying cross-validation---each part is used as the test set, while the rest of the data is used to train the models. The resulting predictions are then used as input to the meta-learner. The meta-learner is trained to make the final prediction based on the predictions of the individual base-learners. This is known as the "winner-takes-all" strategy and allows the ensemble to benefit from the strengths of different ML algorithms and make predictions with better performance than any individual base-learner in the ensemble \citep{Ganaie_2022}.

There is limited research addressing the breadth of base-learners used by ensembles sentiment classification. Augustyniak et al. \citep{augustyniak2014simpler} suggested the use of an ensemble of lexicon-based base-learners as an improvement for using supervised Bag-of-Words (BOW) models in some datasets. Similarly, Saleena et al. \citep{saleena2018ensemble} proposed an ensemble classification system of supervised models, including a Random Forest approach based on BOW, to improve performance. Gifari and Lhaksmana \citep{9702088} evaluated four ML models-- Multinomial Naive Bayes, k-Nearest Neighbors, Logistic Regression, and an ensemble of these models, for SA of movie reviews taken from IMDB, finding that the ensemble model achieved the highest accuracy of 89.40\% among the methods tested. 

All in all, prior attempts for using ensembles for SA were typically limited to base-learners from a single category or from two subsequent (evolution-wise) categories, e.g., lexicon-based and BOW traditional ML \citep{prabowo2009sentiment, 7856499} or BOW traditional ML and neural networks methods with word embedding \citep{mohammadi2021ensemble, KAZMAIER2022115819}. To the best of our knowledge, there were no attempts to use transformers, which are the current state-of-the-art base-learners, alongside weaker methods such as lexicon-based and base-learners of other classes as part of an ensemble, as suggested in the current paper. Furthermore, while existing ensemble construction methods either consider the full set of base-learners or apply pruning technique, our proposed HEC algorithm builds the ensemble from scratch, in a process offering several important advantages, as discussed above.

\section{The HEC Algorithm}

The intrinsic challenge in augmenting the performance of ensembles is rooted in the identification of an optimal subset of base-learners. Given the significant variation in accuracy across base-learners within a heterogeneous ensemble, a straightforward aggregation of the entire set may lead to compromised accuracy. This underscores the imperative for a more nuanced approach to constructing diverse ensembles that surpass the limitations inherent in homogeneous configurations.

In response to this challenge, we introduce the HEC algorithm. HEC aims to overcome the natural tendency of ensemble construction methods to heavily rely on the strongest (accuracy-wise) base-learners in the set available. This may otherwise result in including mostly transformers in the ensemble, rather than effectively leveraging the strengths of a broader spectrum of methodologies, specifically those that are lexicon-based, BOW or word-encoding methods.  As mentioned above, HEC builds the ensemble from scratch, rather than attempting to apply pruning.  This is achieved through a greedy-based selection mechanism, used to identify and incorporate a subset of base-learners that are poised to enhance the overall accuracy of the ensemble from a vast repertoire of base algorithms. 

\begin{algorithm}
\caption{Optimized Hierarchical Ensemble Construction ($L = [X_1, \ldots, X_n]$, $S$)} \label{alg:HEC}
\SetKwInOut{Input}{input}\SetKwInOut{Output}{output}
\Input{$L = [X_1, \ldots, X_n]$: list of base-learners (ordered by accuracy); $S$: maximum initial subset size; $T_{\text{init}}$, $T_{\text{min}}$: initial and minimal temperatures; $coolingRate$: cooling rate} 
\Output{$K$: the suggested ensemble}
\BlankLine
$bestScore \gets 0$;\\
$K \gets$ empty sequence;\\
\BlankLine
\ForEach{subset $M\subseteq L$, $|M|\leq S$}{ \label{alg:BeginLoop}
    $T \gets T_{\text{init}}$;\\
    $previousScore \gets$ Evaluate initial ensemble $M$;\\ \label{alg:Evaluate1}
    $L' \gets L - M$ (keeping the original order of L) \\
    \For{$j = 0$ \KwTo $(n - |M|)$}{ \label{alg:InternalLoop}
        $X_j = L'[j]$; \\ 
        $validationScore \gets$ Evaluate $M \cup \{X_j\}$;\\ \label{alg:AddingBaseLearner}  
        $\Delta E \gets validationScore - previousScore$;\\ ensemble.\\
        \If{$\Delta E \geq 0$ or $\exp(\frac{\Delta E}{T}) > \text{Random}(0, 1)$}{ \label{alg:checkBegin}
            $M \gets M \cup \{X_j\}$;\\
            $previousScore \gets validationScore$;\\
        }  \label{alg:checkEnd}
        $T \gets T \times coolingRate$;\\  \label{alg:cooling} 
        \If{$T \leq T_{\text{min}}$}{
            break; 
        }
    } \label{alg:InternalLoopEnd}
    \If{$previousScore > bestScore$}{  \label{alg:CandidateUpdateBegin}
        $bestScore \gets previousScore$;\\
        $K \gets M$;\\
    }  \label{alg:CandidateUpdateEnd}
} \label{alg:EndLoop}
\Return{$K$}
\end{algorithm}

The logic of HEC is given in Algorithm  \ref{alg:HEC}. HEC examines the problem at hand as a local search problem, with the objective of iteratively adding base-learners as long as such additions improve accuracy, using simulated annealing. The algorithm receives as an input a list of potential base-learners $L = [X_1, ..., X_n]$ that can be used for the ensemble. The list is ordered according to the base-learners' accuracy when applied on the validation set (descending). The construction of the ensemble is carried out by beginning with an initial small subset of base-learners $M$ and expanding it (Steps \ref{alg:BeginLoop}-\ref{alg:EndLoop}). The initial subset is of maximum size $S$, and the algorithm considers all possible initial sets based on the complete list $L$ (Step \ref{alg:BeginLoop}). The value of $S$ is an input to the algorithm, and depends on the availability of computational resources. Naturally, an increase in the value $S$ can potentially improve the ensemble eventually derived and never worsen it.  For each initial subset $M$, the algorithm attempts to add the remaining base-learners in $L$ sequentially, according to their initial order in $L$, which, as explained above, is set by their accuracy (Steps \ref{alg:InternalLoop}-\ref{alg:InternalLoopEnd}). The decision to add an additional base-learner is non-trivial, as a deterioration in accuracy due to adding it can be attributed to encountering a local minimum, and the subsequent addition of additional base-learners could contribute to an enhancement in the ensemble's performance. 

The algorithm therefore employs an exploratory approach, incorporating elements of simulated annealing to enhance its search capability. The next base-learner in line, $X_i$, is initially evaluated in conjunction with the current ensemble $M$ (Step \ref{alg:AddingBaseLearner}), and its marginal added accuracy obtained from adding it is derived (captured by $\Delta E$). The decision on whether to add $X_i$ is based on this latter measure: if the  marginal added accuracy  is positive, or if it is negative but the decrease in accuracy is within a certain probabilistic threshold then the base-learner is added (Step \ref{alg:checkEnd}). Consequently, while the process greedily evaluates the contribution of base-learners according to their accuracy as reflected by their ordering in $L$, it still allows for the acceptance of base-learners that do not immediately improve performance. The threshold for handling the latter case is dynamic and decreases as the number of base-learners already added increases. It is managed by the parameters $T_{\text{init}}$, $T_{\text{min}}$ and $coolingRate$, which are received as an input to the algorithm. The first, $T_{\text{init}}$, is the initial threshold to be used. The second, $T_{\text{min}}$, is the minimum threshold, that upon reaching it the process terminates. The last parameter, $coolingRate$, is the rate by which we exhaust the threshold $T$ as the process progresses.  This temperature-controlled probability allows the algorithm to potentially escape local minima. Finally, the accuracy of the ensemble constructed based on the current initial subset $M$ is compared with the accuracy of the best ensemble constructed so far, and if found to be greater than the current ensemble then it is set as the new best (Steps \ref{alg:CandidateUpdateBegin}-\ref{alg:CandidateUpdateEnd}.

The computational complexity of HEC is $\mathcal{O}\left(\sum_{i=0}^{S} \binom{n}{i} \cdot n \cdot E\right)$, where $S$ is the initial subset size, $n$ is the number of base-learners available, and $\mathcal{O}(E)$ is the complexity of the evaluation function. The term $\sum_{i=0}^{S} \binom{n}{i}$ applies to all seeds (subsets of size $S$ and below that the algorithm begins with. While this latter term grows exponentially, for moderate $S$ values the number is relatively reasonable, e.g., in our evaluation we use $S=3$, and $n=75$(!) which maps to $70376$ possible seeds. For each initial seed the number of iterations needed for the simulated annealing process to reach convergence is bounded by \(\mathcal{O}(n)\), and a single evaluation of complexity $E$ takes place for each such iteration.

We emphasize that the algorithm is agnostic to the aggregation function to be used for the ensemble. The latter can be set to any preferred method, with any desired parameters, and the only adaptation required is the use of this method in Steps \ref{alg:Evaluate1} and \ref{alg:AddingBaseLearner} where the ensemble evaluation takes place. Specifically in our experiments we use WMV (see more details on that in Section \ref{Ensemble Construction Methods}), though any of the other evaluated aggregation functions (Stacking,  Shapley classifier and Bayesian Network) could have been used, as well as other methods suggested in prior work.

\section{Evaluation}
In this section we demonstrate the benefit of using the full set of base-learners, even the less accurate sentiment algorithms, as candidates for inclusion in the ensemble.  We show that this approach yields improvements of ensembles with transformers alone. We also show the success of the HEC algorithm compared to other possible ensemble construction and aggregation algorithms.  

\subsection{Ensemble Construction and Aggregation Methods}
\label{Ensemble Construction Methods}
We employ four established aggregation methods for ensemble aggregation: Bayesian Networks \citep{cordella2013weighted}, WMV \citep{rs13193945}, Stacking \citep{mohammadi2021ensemble}, and Shapley values \citep{rozemberczki2021shapley}. In addition we apply a random ensemble selection as an additional benchmark.

\paragraph{Weighted Majority Voting (WMV):} WMV involves weighting each base-learner's contribution according to its accuracy on a validation dataset. WMV then combines the predictions of the base-learners through a weighted sum of their individual forecasts. The implementation used for WMV adopts a two-step weight assignment method, as suggested by \citep{rs13193945}: First, the weight of each base models class $k\in{1,...,4}$ (in our case: lexicon-based, bag-of-word traditional ML, neural network models with word embedding, and transformer models), denoted $w_k$, is determined based on its average accuracy:
\begin{equation} 
    W_k = \frac{\text{Accuracy}_k}{\sum_{l=1}^4 \text{Accuracy}_l}
\end{equation}
where \(\text{Accuracy}_l\) is the average accuracy of class \(l\). Next, a weight is assigned to each base-learner within its respective class. The weight \(W_{i,k}\) for each base-learner \(i\) in class \(k\) is calculated as the product of its accuracy and the weight of its class, normalized to ensure that the sum of all weights equals 1:
\begin{equation} 
    W_{i,k} = \frac{\text{Accuracy}_{i,k} \times W_k}{\sum_{j=1}^{n_k} \text{Accuracy}_{j,k} \times W_k}
\end{equation}
where \(\text{Accuracy}_{i,k}\) is the accuracy of base-learner \(i\) for class \(k\), and \(n_k\) is the number of base-learners of class \(k\).

The predicted output of the ensemble, \(WMV(x)\), is calculated as the weighted sum of the predictions made by each base-learner:
\begin{equation}
    WMV(x) = \sum_{k=1}^4 \sum_{i=1}^{n_k} W_{i,k} f_{i,k}(x)
\end{equation}
where \(f_{i,k}(x)\) is the prediction of the \(i\)-th base-learner of class \(k\) for input \(x\), and \(W_{i,k}\) is its corresponding weight.

\paragraph{Stacking:} 
The main principle of stacking involves training secondary meta-learners to make a final prediction based on the outputs of the base-learners in the ensemble. Our implementation of Stacking uses Logistic Regression and Random Forest as meta-learners, following Mohammadi and Shaverizade \citep{mohammadi2021ensemble}. These meta-learners were trained using the predictions of the base-learners on a validation set.

\paragraph{The Shapley Value classifier:} 
The main principle of the Shapley Value classifier is to quantify the contribution of each base-learner to the ensemble's overall performance and assign importance accordingly \citep{rozemberczki2021shapley}. This strategy is based on the observation that complicated classifiers are often liable for accurate and inaccurate classification decisions. To evaluate the Shapley Value classifier, we employ three primary suggested implementations found in prior work (Expected Marginal Contribution Approximation \citep{FATIMA20081673}, Multilinear Extension \citep{14e16570-955c-30f7-bcc4-7f2f0ad943cf} and Monte Carlo Permutation Sampling \citep{maleki2014bounding}). 

\paragraph{Bayesian Network:}
Bayesian Networks can be used to combine predictions from multiple base-learners by treating the outputs of these models as evidence in a Bayesian Network. This network then infers the most probable outcome considering the dependencies and uncertainties modeled between the inputs. This method can capture complex relationships and uncertainties among model predictions, leading to a more informed final prediction. Our use of Bayesian Networks as a benchmark for the evaluation follows the implementation of Cordella et al. \citep{cordella2013weighted}.


\paragraph{Random Selection Ensemble:}
As an additional baseline for the comparative analysis we implement a random selection ensemble construction strategy. This approach involves the stochastic selection of base-learners for the ensemble, aggregated by a simple majority voting. Specifically, each base-learner is picked with a 0.5 probability (applying a different randomly picked seed for each selection). This was repeated five times and the highest performing ensemble over the validation set was picked. 

\paragraph{HEC:}
We use an initial subset size of up to three base-learners ($S = 3$). This size strikes a balance between computational resource demands and the potential for marginal improvement in performance. We employed weighted voting for the aggregation function to evaluate the base-learner sequences primarily due to its intuitiveness. The implementation is described above. 
The parameters used for the simulated annealing component are an initial temperature of 0.8, a cooling rate of 0.8 and a minimum threshold of 0.1. These values were selected based on a conservative strategy that emphasizes taking smaller steps to increase the likelihood of exploring larger occasional deviations from the optimal path, as suggested by Weyland \citep{Weyland_Dennis}.

Note that for most of the above mentioned methods pruning is implicitly inherent in their design. For example, a base-learner can be excluded in WMV by assigning it a zero weight. Similarly, the Stacking meta-learner can ignore the input received from a base-learner that needs to be excluded, and the Shapley values classifier can include base-learners as well by assigning adversarial models a low valuation score that barely affects the ensemble. Therefore we do not apply any external pruning techniques, but rather intentionally enable the methods to use the complete set of base-learners, allowing complete flexibility in deciding the extent of reliance on each available base-learner.

\subsection{Base-Learners} 
We use seventy five base-learners (including variations) that the different ensemble construction methods can fully rely on or pick a subset of. These methods can be mapped to the four main categories mentioned earlier in the paper: three Lexicon-based sentiment appraches (3 out of 75), traditional ML with several BOW / W2V configurations (60 out of 75), convolutional neural networks with two different word embeddings (3 out of 75), and transformer-oriented base-learners (9 out of 75). In the following paragraphs we review the four categories and detail the methods used within each one. 


\paragraph{Lexicon-Based Methods:} These approaches assume that the semantic orientation of a text is predominantly determined by the polarity of the words and phrases contained within. This perspective is based on the belief that the content words within a text are strongly correlated with its overall semantic orientation \citep{Turney2002}, hence the sentiment of a text can be determined by some aggregation of the polarity (typically -1 for negative polarity and +1 for positive, with additional variations for score) of the different words in it. For the evaluation we consider three lexicon-based base-learners: VADER, Pattern, and AFINN \citep{hutto2014vader,deSmedt2012Pattern,nielsen2011new}. VADER (Valence Aware Dictionary and Sentiment Reasoner) is one of English's most highly regarded polarity lexicons. VADER lexicon has been shown to perform exceptionally well in the domain of social media. In particular, a correlation coefficient analysis revealed that it performs comparably to individual human raters in terms of its ability to align with ground truth sentiments as determined by an aggregated group of 20 human raters \citep{hutto2014vader}. Pattern \citep{deSmedt2012Pattern} is a web mining module for Python, which uses WordNet English adjectives present in a given text. Finally, AFINN \citep{nielsen2011new} is a widely used lexicon 
which assigns a score to a given word on a scale of -5 to 5 indicating the polarity of the word. We follow Vashishtha and Seba \citep{vashishtha2022neuro} and classify the texts according to the sum of the binary polarity of the words composing them. All three base-learners were implemented based on their tools and libraries available on GitHub \citep{VaderSentiment,Pattern,Afinn}.

\paragraph{Bag-of-word Approaches} 
These approaches use a Bag-of-word (BOW) text encoding to extract a set of features from the training data, and use traditional ML classifiers such as Naive Bayes (NB), Support Vector Machine (SVM), Logistic Regression (LR), or Random Forest (RF) to train the base-learner on the sentiment labels. Once trained, the classifier can be used to predict the sentiment orientation of an unlabeled sample. In this study, we consider the above four classifiers (NB, SVM, LR and RF). For each, we produce 15 variations, differing in the embedding method, the tokenizer and the N-gram used. The first 12 variations are BOW-based, with three embedding methods, two tokenizers and two N-gram options (see Table \ref{tab:configurations}). In addition to these, we use three vector representation embedding (W2V (300D) freezed / trainable, GLOVE(200D)) yielding additional three variants for each classifier. These options yield a total of sixty distinct base-learners.

\begin{table}[ht]
\centering
\footnotesize
\begin{tabular}{lll}
\toprule
\textbf{Configuration Component} & \textbf{Options} & \textbf{Details} \\
\midrule
\multirow{3}{*}{Embedding (BOW)} & Binary & - \\
 & Frequency & - \\
 & TF-IDF & - \\
\midrule
\multirow{2}{*}{Tokenization/Stemming} & SkLearn default & - \\
 & Porter's + NLTK \citep{mhatre2017dimensionality}& - \\
\midrule
\multirow{2}{*}{N-gram} & Uni-gram & - \\
 & Bi-gram & - \\
\midrule
\multirow{2}{*}{Vector Representation} & W2V (300D) \citep{mikolov2013efficient}& Trainable / Frozen \\
 & GloVe (200D) \citep{pennington-etal-2014-glove} & \\
\midrule
\multicolumn{2}{l}{\textbf{Total Configurations for BOW:}} & $3 \times 2 \times 2 = 12$ \\
\multicolumn{2}{l}{\textbf{Total Configurations for Vector Representations:}} & $3$ \\
\multicolumn{2}{l}{\textbf{Total Combined Configurations:}} & $(12 \text{ (BOW)} + 3 \text{ (Vector)}) \times 4 \text{ (Classifiers)} = 60$ \\
\bottomrule
\end{tabular}
\caption{Summary of traditional ML configurations of base-learners.}
\label{tab:configurations}
\end{table}

\paragraph{Convolutional Neural Networks (CNN):} While Cnns are typically used for image processing, these methods have gained significant attention as means for SA. A prominent SA CNN model was proposed by Kim \citep{DBLP:journals/corr/Kim14f}, assessing a CNN model constructed atop a pretrained Word2vec embedding to perform sentiment classification at the sentence level. This model demonstrated superior performance compared to alternative methods, substantiating the efficacy of pretrained word embedding as useful features for NLP tasks in conjunction with deep learning \citep{CHEN2017221}.

We use the CNN architecture suggested by Kim \citep{DBLP:journals/corr/Kim14f}. Particularly, we implement a non-Static model featuring a trainable pretrained embedding layer and three convolutional layers with filters of sizes 3, 4, and 5, followed by a max-pooling over time layer. We implement three CNN architectures: two using Word2Vec (trainable pretrained / non trainable-pretrained) (Google 300D) embeddings and the last using GLOVE embeddings \citep{Word2Vec,Glove-PenningtonSM14}. We trained these models for 25 epochs, evaluating each epoch on the development set, and selected the model corresponding to the epoch with the highest accuracy score, as recommended by Kim. Differing from the original paper, we utilize the Adam optimizer with a learning rate of 0.002 and a beta value of 0.9, following Britz and Denny's implementation \citep{Britz2015}.

\paragraph{Transformers:} 
Transformers are applied to a wide range of text classification problems, including binary sentiment classification. 
Pretrained transformer models are used for SA tasks in several manners: as text classifiers (by adding additional layers connecting the output of the transformer model with the sentiment labels), for evaluation (by calculating distances between the unseen texts and texts of which the sentiments are already known), and as zero-shot models that can evaluate the relatedness of some word (category, narrative, etc.) with the text \citep{mishev2020evaluation}.

We consider nine well established transformer-based base-learners: BERT\citep{devlin2018bert}, roBERTa \citep{liu2019roberta}, alBERT \citep{lan2020albert}, DistilBERT \citep{sanh2020distilbert}, ELECTRA\citep{clark2020electra}, XLNET \citep{yang2020xlnet}, and SiEBERT \citep{HARTMANN202375}, which were all fine-tuned for five epochs on each data-set (see summary in Table \ref{Transformers:configurations}). In addition, we use a zero-shot pretrained transformer based on BART \citep{lewis2019bart} and a few-shot transformer based on the SetFIT architecture \citep{tunstall2022efficient}, which was trained for 10 epochs and 100 samples for each dataset. The specific configuration of each different transformer base-learner follows the one suggested by Pipalia et al and Mishev et al \citep{9337081,9142175}. 

\begin{table}
\scriptsize
\centering
\begin{tabular}{lcc}
\toprule
Model & Model Info & Fine-Tuning \\
\midrule
BERT & Base-uncased & 5 epochs \\
XLNet & Base-cased & 5 epochs \\
RoBERTa & Base & 5 epochs \\
DistilBERT & Base-uncased & 5 epochs \\
ALBERT & Base-v1 & 5 epochs \\
ELECTRA & Base & 5 epochs \\
SiEBERT & Base & 5 epochs \\
BART & Large-mnli & Zeroshot \\
SetFit & MPNET (FewShot) & 10 epochs \\
\bottomrule
\end{tabular}
\caption{Transformer models and fine-tuning details}
\label{Transformers:configurations}
\end{table}

\subsection{Datasets}
We evaluate the different ensembles using a diverse selection of eight datasets from canonical sentiment problems, the details of which are summarized in Table \ref{tab:case_study_data_sets}. 
 These datasets were shuffled and divided into training, validation (development), and test sets. The validation datasets were used to calculate the relative weights of the various base-learners within each ensemble technique. 
In order to facilitate replication of these results, all datasets employed in this study can be accessed on the authors' Hugging Face Hub.\footnote{Huggingface Link: \citep{huggingface_pig4431}} In some of the datasets, only a portion of the data was selected due to computational constraints. Particularly, in the IMDB \citep{maas-EtAl:2011:ACL-HLT2011}, Amazon Polarity, and YELP datasets \citep{maas-EtAl:2011:ACL-HLT2011,NIPS2015_250cf8b5}, A random selection of 50,000 reviews from the corpus was made and divided into three sets: 25,000 for training, 12,500 for validation, and 12,500 for testing. Additionally, for the YELP dataset, reviews with 1 or 2 stars were labeled as negative, and those with 4 or 5 stars were labeled as positive, with reviews bearing 3 stars being filtered out. 

\begin{table*}[htbp]
\scriptsize
\caption{A summary of the eight datasets used for evaluation, specifying the average sentence length in words.}
\label{tab:case_study_data_sets}
\begin{tabularx}{\textwidth}{lccccX}
\toprule
    Dataset & Domain & AVG Sentence Length (Words) & \multicolumn{3}{c}{Quantity of data} \\
\cmidrule{4-6}
& & & Training & Validation & Test \\
\midrule
Yelp (YLP) \citep{yelp_dataset} & Business Reviews & 131.14 & 25,000 & 5,000 & 5,000\\
IMDB \citep{imdb_dataset} & Movie Reviews & 233.79 & 25,000 & 12,500 & 12,500 \\
SST2 \citep{sst_dataset} & Movie Reviews & 19.30 & 67,349 & 872 & 1,821 \\
Tweet Sentiment Extraction (TSE) \citep{tse_dataset} & Social Media & 13.28 & 14,200 & 2,100 & 2,100\\
Twitter US Airline Sentiment (TUF) \citep{tuf_dataset} & Social Media & 18.50 & 8,000 & 1,100 & 2,300\\
SemEval-2016 TweetEval(Sentiment) (TE) \citep{te_dataset} & Social Media & 19.45 & 20,000 & 6,300 & 6,300\\
Rotten Tomato (RTM) \citep{rtm_dataset} & Movie Reviews & 20.99 & 8,500 & 1,000 & 1,000\\
Customer Reviews (CR) \citep{cr_dataset} & Product Reviews & 19.95 & 2,200 & 750 & 750\\
\bottomrule
\end{tabularx}
\end{table*}

\subsection{Comparison to GPT-4}
To evaluate the performance of GPT-4 for the sentiment task, we used the following prompt, as suggested by Kheiri and Karimi \citep{kheiri2023sentimentgpt}: 
\begin{quote}
\textit{"As a social scientist, your task is to analyze the sentiment of a series of statements extracted from various outlets. Please assign a sentiment score from 0 to 1 for each statement, where 0 signifies negative sentiment, and 1 corresponds to positive sentiment. In situations where the sentiment is difficult to definitively classify, please provide your best estimation of the sentiment score."}
\end{quote}

The outputs produced by the GPT-4 language model are then processed and labeled into negative if the score is below 0.5 and positive otherwise, a similar process to an argmax operation in a neural network.


\section{Results}

For clarity, we limit this section to contrasting HEC with WMV and Stacking. WMV and Stacking are selected for their well-established reputation in enhancing deep learning ensemble performance \citep{MOHAMMED2023757}. The comparison with the other evaluated methods, yielding largely similar qualitative conclusions and is provided in the appendix.

Table \ref{results:main} summarizes the performance (average accuracy) obtained with the primary methods tested for the different datasets used and the average performance cross-datasets. The first column provides the performance of HEC. The next three columns provide the performance of the two leading ensemble construction methods (WMV and Stacking), and random ensemble construction, when the set of methods available includes the full set of the 75 base-learners. The next three columns provide similar data, for the case where only transformer-based base-learners are available for the ensemble construction methods. The next column provides the performance of the GPT-4 LLM. Finally, the last four columns depict the performance of the four top transformers when used as a stand-alone and not part of an ensemble. These four base-learners are those that scored highest for at least one dataset.

As part of the results analysis we also report the extent/portion of the gap between the top-performing individual base-learner and a perfect classifier (100\% accuracy) that the evaluated ensemble managed to eliminate. This is important mainly because, as discussed earlier, current transformers are at a breech of perfection in predicting SA. The latter measure thus reflects what portion of the remaining gap to 100\% accuracy can be eliminated through the use of the investigated ensemble. Formally, given a base-learner or ensemble $A$, the eliminated portion of the gap to perfection that an ensemble $B$ manages to eliminate is given by: $(Acc_{B}-Acc_{A})/(1-A_{A})$, where $Acc_{B}$ and $Acc_{A}$ are the accuracies of methods $B$ and $A$, respectively.

\begin{table*}[]
\scriptsize
\begin{tabular}{|l|c|ccc|ccc|c|cccc|}
\hline
\multicolumn{1}{|l|}{} &      & \multicolumn{3}{c|}{Ensemble - All} & \multicolumn{3}{c|}{Ensemble - Transformers}                                 & LLM     & \multicolumn{4}{c|}{Non-Ensemble}                                                                 \\ \cline{3-13}
Dataset                   & HEC     & WMV           & Stack         & Random       & WMV                         & Stack                       & Random  & GPT4    & roBERTa                     & ELECTRA                     & XLNet                       & SiEBERT \\ \hline
TSE                    & \textbf{96.53\%} & 95.17\%       & 95.50\%       & 92.48\%      & 95.00\%                     & 94.24\%                     & 94.95\% & 91.0\%  & 94.62\%                     & 94.58\%                     & 93.58\%                     & 88.62\% \\
SST2                   & \textbf{96.37\%} & 95.17\%       & 95.17\%       & 92.48\%      & 95.16\%                     & 95.22\%                     & 94.95\% & 94.4\%  & 94.62\%                     & 94.45\%                     & 92.97\%                     & 94.50\% \\
CR                     & \textbf{95.61\%} & 94.56\%       & 94.95\%       & 87.92\%      & 94.68\%                     & 94.15\%                     & 94.69\% & 92.7\%  & 93.62\%                     & 93.89\%                     & 93.62\%                     & 87.51\% \\
TE                     & \textbf{94.29\%} & 93.21\%       & 93.65\%       & 88.01\%      & 93.72\%                     & 93.69\%                     & 93.35\% & 94.8\%  & 93.65\%                     & 90.70\%                     & 90.90\%                     & 89.85\% \\
IMDB                   & \textbf{96.33\%} & 96.29\%       & 96.21\%       & 94.35\%      & 95.76\%                     & 96.05\%                     & 95.94\% & 94.7\%  & 95.80\%                     & 95.66\%                     & 92.04\%                     & 95.67\% \\
TUF                    & 96.62\% & 96.28\%       & 96.02\%       & 95.06\%      & 96.10\%                     & 96.36\%                     & 96.28\% & \textbf{97.2\%}  & 95.15\%                     & 94.02\%                     & 96.10\%                     & 95.93\% \\
YLP                    & \textbf{97.86\%} & 97.30\%       & 97.76\%       & 95.68\%      & 97.50\%                     & 97.73\%                     & 95.94\% & 97.8\%  & 97.36\%                     & 94.90\%                     & 96.10\%                     & 97.05\% \\
RTM                    & \textbf{92.03\%} & 92.03\%       & 90.53\%       & 85.83\%      & 90.42\%                     & 89.39\%                     & 89.68\% & 89.2\%  & 88.36\%                     & 90.61\%                     & 88.46\%                     & 92.03\% \\ \hline
Average                & \textbf{95.71\%} & 95.00\%       & 94.97\%       & 91.48\%      & 94.79\%                     & 94.60\%                     & 94.47\% & 93.97\% & 94.15\%                     & 93.60\%                     & 92.97\%                     & 92.64\% \\ \hline
\end{tabular}
\caption{Accuracy comparison across various methods for selected datasets. This table presents the performance of ensemble and non-ensemble models, including a comprehensive ensemble of all seventy-five base-learners ("Ensemble - All"), and a focused ensemble of transformer-based learners only ("Ensemble - Transformers"). For individual, non-ensemble approaches, performance metrics of leading transformer models are displayed. Abbreviations: HEC (Hierarchical Ensemble Construction), WMV (Weighted Majority Voting), Stack (Stacking), LLM (Large Language Model, specifically GPT-4), individual transformer models. The table outlines the accuracy percentages for each method across different datasets, providing a direct comparison to assess the efficacy of ensemble versus single-model strategies.}
\label{results:main}
\end{table*}

\subsection{Ensemble Methods vs. Individual Base-Learners}

We observe from Table \ref{results:main} that the use of ensembles indeed results in better performance compared to using any of the individual base-learners exclusively---both top evaluated ensemble construction methods (WMV and Stacking) scored better on accuracy, on average, cross datasets, compared to the top four evaluated transformer base-learners. This holds both when the ensembles are based solely on transformers and when relying on the full set of 75 base-learners.  The phenomenon is generally consistent cross datasets---each of the ensembles produced with WMV and Stacking (both when relying on transformers and on the full set of base-learners) performed better than the top base transformer learner in six out of the eight datasets. Random ensembles are found to be highly effective when relying on transformers base-learners, resulting in average accuracy (cross datasets) which is surprisingly close to the results of the other two methods. This latter phenomenon is consistent cross the individual datasets, suggesting that any random aggregation of transformer-based predictions is likely to be quite precise, as the ensemble produced contains sufficient correct individual predictions to cover for an occasion stumbling of one of the base-learners. This, however, does not hold with Random Ensembles constructed over the full set of the 75 base-learners, suggesting that in this case the set used contains a non-negligible number of methods that predict a false outcome.  Consequently, when enabling also the use of non-transformer base-learners, especially lexicon-based and bag-of-word traditional ML, the proper selection of a subset of base-learners to be included in the ensemble becomes crucial to its success.

\subsection{Improving Non-Hec Ensembles with all base-learners}

We observe from Table \ref{results:main} that indeed extending the set of base-learners that can be used with existing methods to include non-transformer methods, obtains, on average, better accuracy. The improvement achieved, however, is somehow moderate---the average accuracy with WMV and Stacking increased from 94.79\% and 94.6\% when relying exclusively on transformers to 95\% and 94.97\% when extending the set of base-learners to other methods, respectively.  Furthermore, this slight improvement is not consistent across datasets, and in three out of the eight datasets both WMV and Stacking, when relying on transformers, performed better than when having the option to rely on the entire range of base-learners. 

All in all, enabling WMV and Stacking using the extended set of base-learners accounts to reducing 14.59\% and 14.12\% of the gap between the top-performing base-learner (roBERTa) and perfection, whereas when these methods rely exclusively on transformers they manage to reduce only 11.02\% and 7.80\% of that remaining gap. Thus, we note that the improvement resulting from enabling the use of the full set of base-learners is quite slight. In fact, even using an ensemble exclusively constructed from random transformers managed at times to yield better accuracy than with the ensembles constructed by WMV and Stacking that use the full set of base-learners (e.g., for CR and TE with WMV and for TUF when using Stacking). In the following paragraphs we present the results of HEC, demonstrating that when relying on HEC, the use of a mix of model types significantly outperforms traditional transformer-only ensembles.

\subsection{HEC Performance}

We observe from Table \ref{results:main} that HEC's mean accuracy (cross datasets) is 95.71\%. This is far better than WMV (94.79\%) and Stacking when using the full set of base-learners. In fact, HEC offers an improvement which is 4.4 and 3 times greater than the improvement in the WMV and Stacking ensembles respectively make when these two methods when they use only transformers.

Furthermore, even when each specific dataset was considered individually, we find that none of the other methods (individual transformers, WMV/Stacking/Random ensembles based on transformers, and WMV/Stacking/Random ensembles based on the full set) manage to outperform HEC (see also Figure \ref{fig:hec_performance}). This latter observation is particularly impressive, given that the WMV and Stacking ensembles that made use of the full set of base-learners failed to improve the performance of their all-transformers ensemble variant for some of the datasets. 

HEC managed to reduce 26.61\% of the gap between the performance of the top-performing base-learner (roBERTa) and a perfect classifier (compared to 11.02\% and 7.80\% that WMV and Stacking based on transformers, and 14.59\% and 14.12\% that these methods when relying on the entire set of base-learners managed to bridge, respectively).\footnote{The top-performing base-learner is 5.85\% close to 100\% accuracy, and with HEC we managed to reduce that gap to 4.29\%.}

\begin{figure}[htbp]
\centering
\includegraphics[width=0.7\textwidth]{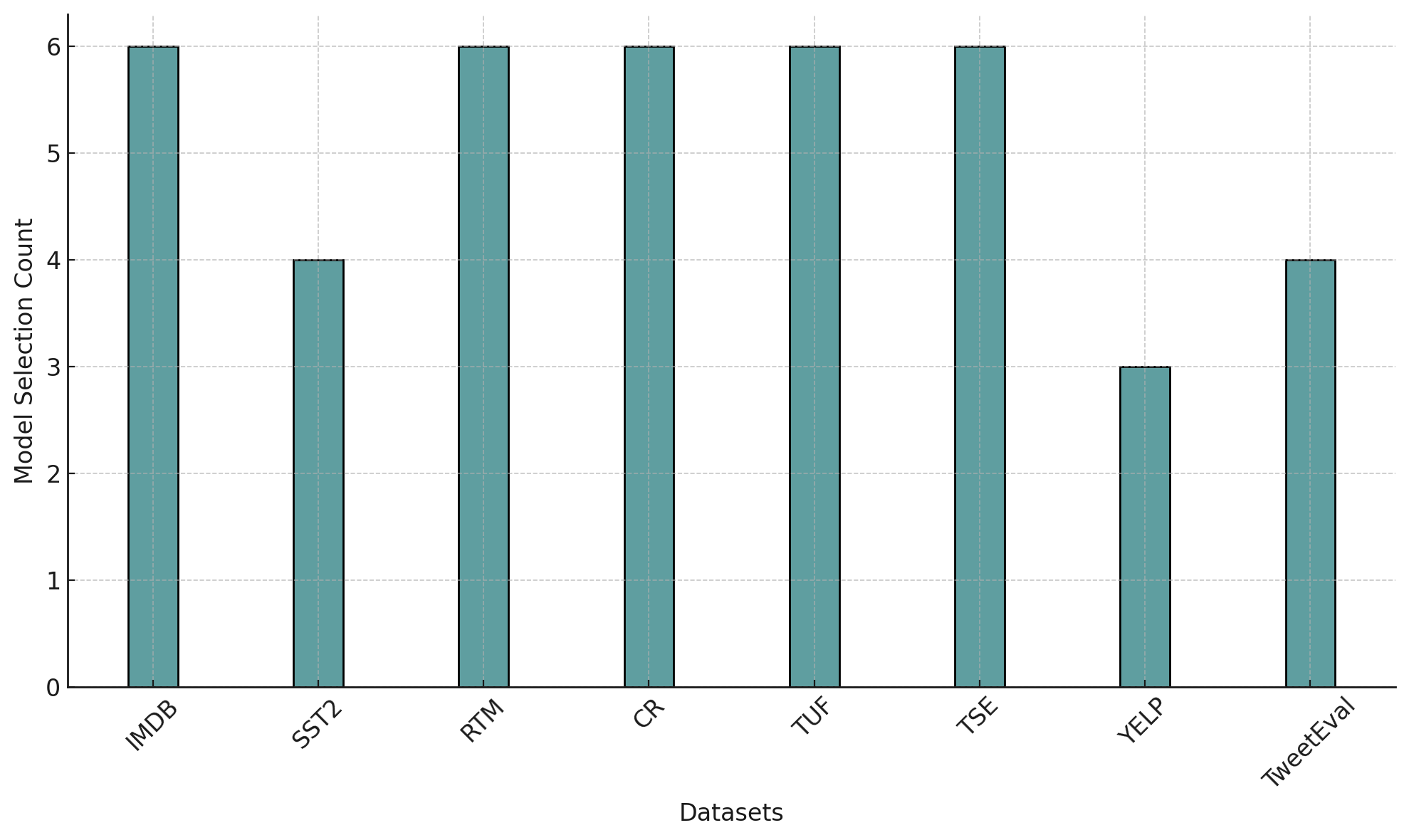}
\caption[Model Inclusion Count in HEC Ensemble]{Number of base-learners included in the HEC ensemble for the different datasets.}
\label{fig:inclusion-count-hec}
\end{figure}

\begin{figure}[htbp]
\centering
\includegraphics[width=0.7\textwidth]{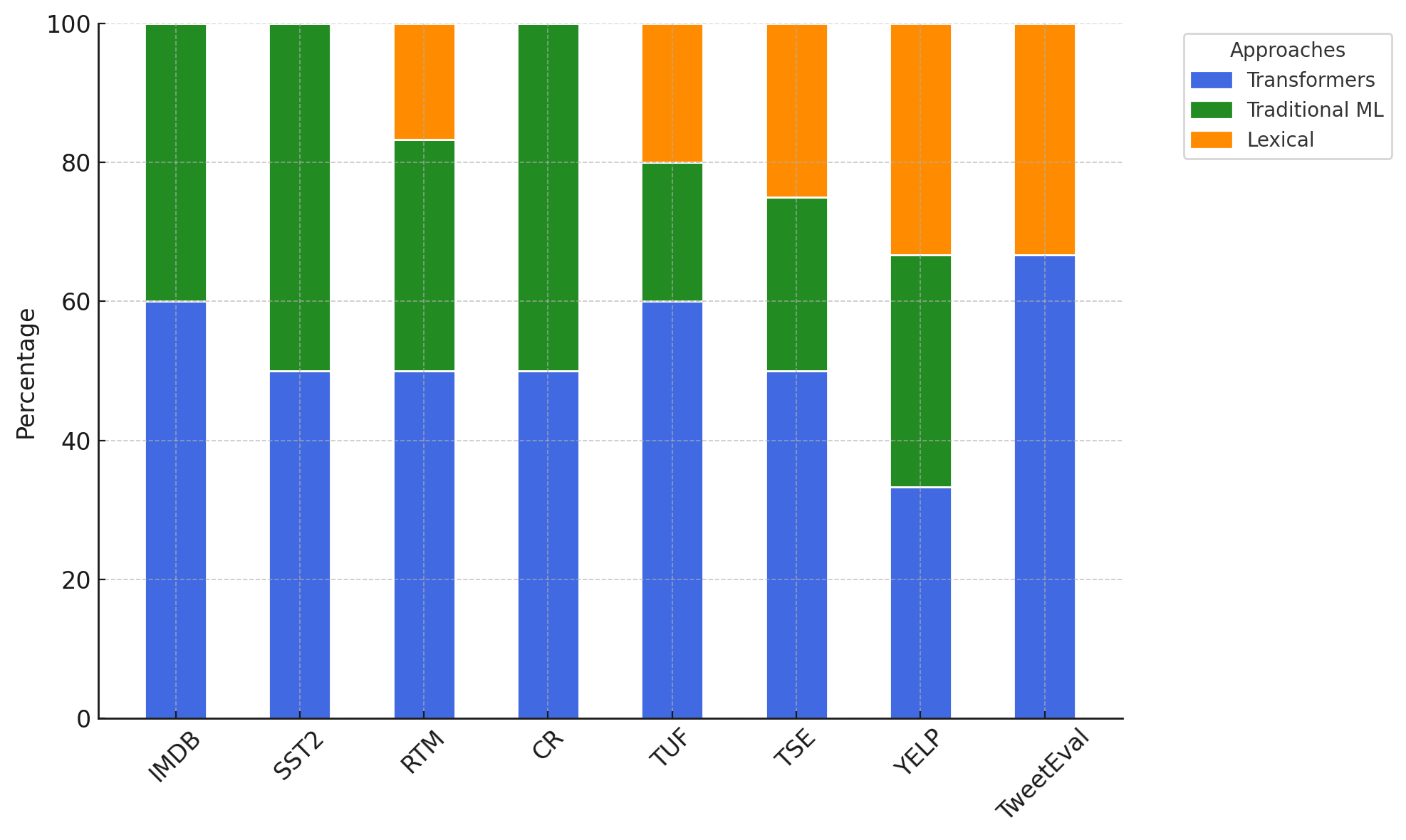}
\caption[HEC Model Inclusion Histogram]{Percentage inclusion of models by learning algorithm in the HEC ensemble.}
\label{fig:hec-histogram}
\end{figure}

\begin{figure}[htbp]
\centering
\includegraphics[width=0.7\textwidth]{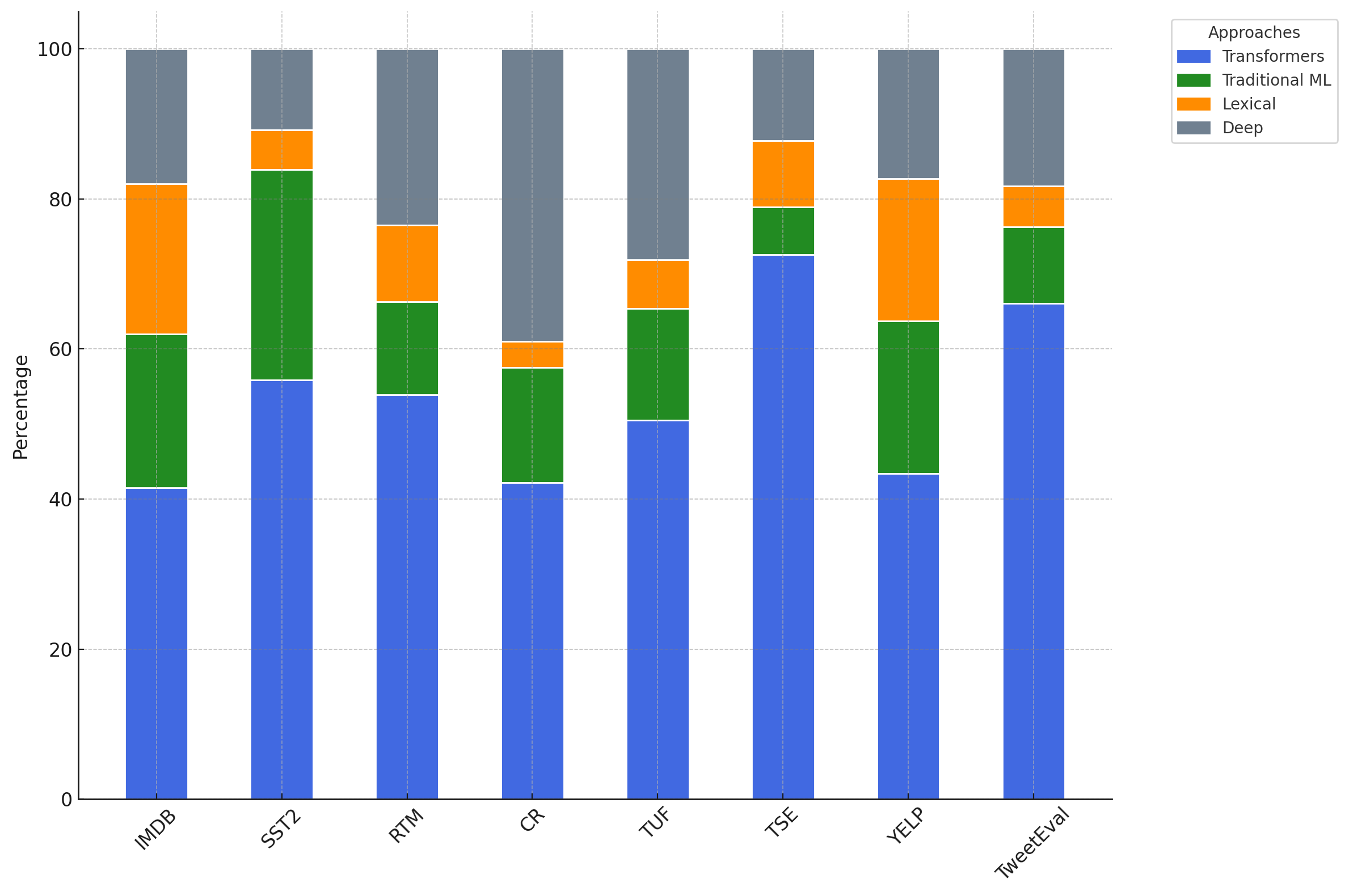}
\caption[WMV Model Inclusion Histogram]{Percentage inclusion of models by learning algorithm in the WMV ensemble.}
\label{fig:wmv-histogram}
\end{figure}

\begin{figure}[htbp]
\centering
\includegraphics[width=0.7\textwidth]{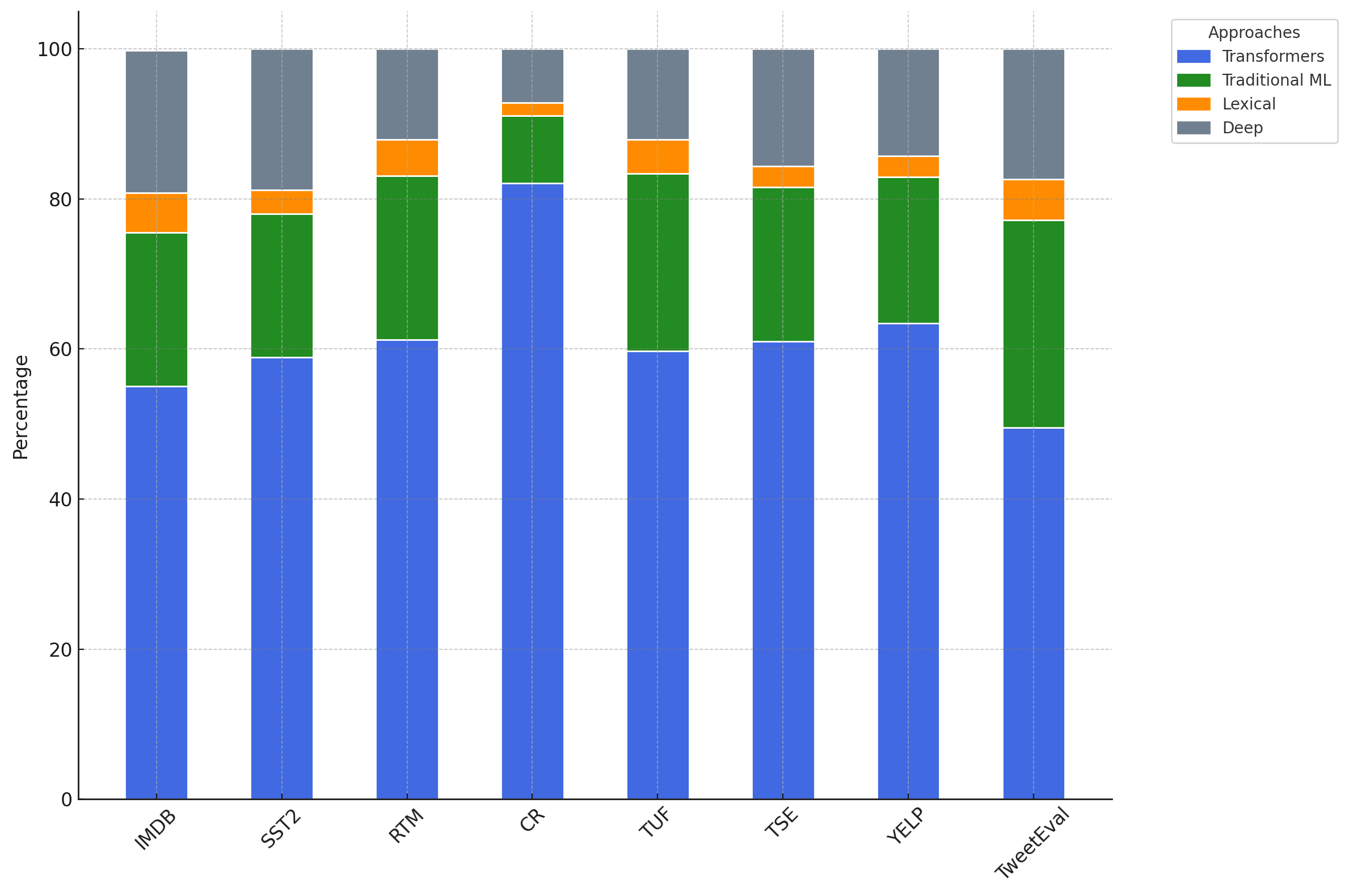}
\caption[Stacking Feature Importance Histogram]{Percentage of models after summing feature importance across each learning algorithm.}
\label{fig:stacking-histogram}
\end{figure}

In order to better understand HEC's success, we present results in Figures \ref{fig:inclusion-count-hec}-\ref{fig:stacking-histogram}. Figure \ref{fig:inclusion-count-hec} depicts the number of base-learners used by HEC for each of the ensembles used for the different datasets. It is complemented by Figure \ref{fig:hec-histogram} which depicts the internal division between the different base-learners according to the four classes used. We observe from Figure \ref{fig:inclusion-count-hec} that even though the input for HEC included 75 base-learners, in all eight datasets HEC produced strikingly small ensembles of only 3-6 base-learners. This strengthen our hypothesis that a large number of base-learners may produce some redundancy which, at times, may negatively influence the aggregation, resulting in predicting the wrong outcome. We observe from Figure \ref{fig:hec-histogram} that HEC completely avoided the use of the neural network base models with word embedding, despite their relatively high accuracy. Instead it used primarily transformer-based and bag-of-word traditional ML base-learners and in some cases also some lexicon-based.  The use of the first type is quite clear, as transformer-based base-learners are the most accurate ones. Alas the two other types, which account for 40-66\% of the base-learners used, are the least accurate ones. The improved performance is thus not the result of improving the accuracy of the individual base-learners used but rather the use of a diverse set of votes. 

The above phenomenon also recurs when analyzing the internal structure of the ensembles produced with WMV and Stacking. Figure \ref{fig:wmv-histogram} depicts the accumulated weight of base-learners of each class, in percentages, when using WMV, for each dataset. We observe from the figure that as was the case for HEC, the produced ensemble relies to a similar extent on transformer-based base-learners. However, while HEC avoids the use of neural network models with word embedding, WMV relies heavily on such methods at the expense of the other two classes. Similarly, Figure \ref{fig:stacking-histogram} depicts the aggregated importance of base-learners of each class in the Stacking ensembles, as reflected by a feature importance analysis (where each base-learner is taken to be a feature in the model). This figure shows a somehow similar pattern, yet the emphasis on transformer-based base-learners is greater compared to HEC and WMV, at the expense of the other classes.

We observe that HEC achieves a standard deviation of $\pm$1.79\% in the performance of the different methods. This indicates that HEC has a relatively consistent performance across various datasets. In comparison to the other methods tested, HEC is second to WMV (when using all base-learners available) which is associated with a standard deviation of $\pm$1.72\%, and the standard deviation of all other methods is substantially greater. 


\begin{table}[ht]
\centering
\begin{tabular}{lcc}
\toprule
\textbf{Method} & \textbf{Configuration/Model} & \textbf{Standard Deviation (\%)} \\
\midrule
Ensemble & All - WMV & 1.72 \\
Ensemble & HEC & 1.79 \\
Ensemble & Transformers - WMV & 2.08 \\
Ensemble & All - Stack & 2.14 \\
Ensemble & Transformers - Random & 2.14 \\
Ensemble & Transformers - Stack & 2.49 \\
Ensemble & All - Random & 3.72 \\
Non-Ensemble & ELECTRA & 1.89 \\
Non-Ensemble & roBERTa & 2.63 \\
Non-Ensemble & XLNet & 2.56 \\
Non-Ensemble & SiEBERT & 3.65 \\
\bottomrule
\end{tabular}
\caption{Standard Deviation (in accuracy) of the different evaluated methods. }
\label{tab:std_deviation}
\end{table}

\subsection{HEC vs. GPT-4}

As can be seen in Table \ref{results:main}, we find that HEC outperforms GPT-4 despite the latter's advanced language understanding and continual learning and adaptation. In fact, all of the ensemble-based methods that are given in the table (except Random with the full set) and even one of the stand-alone transformers (roBERTa) managed to provide better accuracy than GPT-4. Most notable is the fact that even the random transformers-based ensemble performs better than GPT-4. Interestingly, when considering the specific datasets, we find that the performance of GPT-4 highly varies---while in some datasets (e.g., TSE and CR) its performance is way behind HEC, in others (TE and TUF) it actually managed to perform slightly better (see also Figure \ref{fig:hec_performance}). In fact, GPT-4 is the only method among those evaluated in this paper that managed to outperform HEC in some individual dataset.

\section{Discussion and Conclusions}
\label{conclusion}

This research presents a comprehensive examination of ensemble methods in SA (SA), with a specific focus on the novel HEC algorithm. Our findings underscore the superiority of HEC over both traditional ensemble methods and state-of-the-art transformers, including GPT-4, across a diverse set of SA datasets.  Notably, HEC achieves a remarkable mean accuracy of 95.71\% across eight canonical SA datasets, significantly enhancing the performance by effectively bridging the gap toward perfection.

Our results affirm the foundational premise of ensemble learning; that is, a judicious mix of diverse base-learners can substantially mitigate their individual weaknesses. Specifically, HEC's ability to selectively incorporate a broad spectrum of base-learners, from transformers to traditional ML models, underpins its enhanced performance. HEC not only offers superior accuracy but also demonstrates remarkable robustness and consistency across different datasets, as evidenced by its relatively low standard deviation in performance.

It is important to stress that the idea of incorporating in the ensemble lexicon-based, bag-of-word traditional ML and other types of base-learners, alongside transformer-based, and the use of HEC for picking a small, yet highly effective, subset, is general. In this paper we used these ideas for SA, however these can be useful to other NLP classification tasks.  While such evaluation is beyond the scope of this paper, we present in the appendix results of using HEC with fine-tuned transformers for the popular 20Newsgroups dataset, which has 20 categories (rather than two as in our case) and is far more general \citep{lang1995newsweeder}, demonstrating once again a performance improvement compared to existing methods. 

\bibliographystyle{apalike}
\bibliography{references}
\newpage

\appendix
\section*{Appendix}



\section{Performance Comparison Tables}

\begin{table}[ht]
\scriptsize
\begin{tabular}{l|c|cc|cc|cc|cc|cc}
\hline
& & \multicolumn{2}{c|}{WMV} & \multicolumn{2}{c|}{Stack} & \multicolumn{2}{c|}{Shapley} & \multicolumn{2}{c|}{Bayesian} & \multicolumn{2}{c}{Random} \\
Data & HEC & Trans. & All & Trans. & All & Trans. & All & Trans. & All & Trans. & All \\ \hline
TSE & 96.53\% & 95.00\% & 95.17\% & 94.24\% & 95.50\% & 95.19\% & 95.33\% & 94.82\% & 87.70\% & 94.95\% & 92.48\% \\
SST2 & 96.37\% & 95.16\% & 95.17\% & 95.22\% & 95.17\% & 94.21\% & 95.50\% & 95.12\% & 87.70\% & 94.95\% & 92.48\% \\
CR & 95.61\% & 94.68\% & 94.56\% & 94.15\% & 94.95\% & 93.91\% & 94.69\% & 94.60\% & 84.59\% & 94.69\% & 87.92\% \\
TE & 94.29\% & 93.72\% & 93.21\% & 93.69\% & 93.65\% & 94.12\% & 94.00\% & 92.63\% & 83.50\% & 93.35\% & 88.01\% \\
IMDB & 96.33\% & 95.76\% & 96.29\% & 96.05\% & 96.21\% & 94.92\% & 95.82\% & 95.39\% & 91.62\% & 95.94\% & 94.35\% \\
TUF & 96.62\% & 96.10\% & 96.28\% & 96.36\% & 96.02\% & 95.34\% & 96.19\% & 95.89\% & 93.77\% & 96.28\% & 95.06\% \\
YLP & 97.86\% & 97.50\% & 97.30\% & 97.73\% & 97.76\% & 96.12\% & 96.90\% & 97.25\% & 93.82\% & 95.94\% & 95.68\% \\
RTM & 92.03\% & 90.42\% & 92.20\% & 89.39\% & 95.53\% & 90.25\% & 89.96\% & 90.10\% & 83.21\% & 89.68\% & 85.83\% \\
Average & 95.71\% & 94.79\% & 95.00\% & 94.60\% & 95.59\% & 94.25\% & 94.79\% & 94.48\% & 88.24\% & 94.47\% & 91.48\% \\ \hline
\end{tabular}
\caption{Comprehensive comparison of HEC, Transformer-based ensembles, and full model-set ensembles, including Shapley classifier and Bayesian Network.}
\end{table}

\section{20NewsGroups Performance Evaluations}
\begin{table*}[htbp]
\scriptsize
\begin{tabular}{llllllllll}
\toprule
 Data-set   & HEC      & WMV   & Shapley   & Stacking   & Bayesian   & Random (All)   & Random (Trans)   & Non-Ensemble   \\
 \midrule
20NewsGroups        & 89.43\%     & 86.85\%       & 85.43\%     & 87.33\%      & 86.85\%      & 78.34\%          & 86.35\%                   & RoBERTa, 85.97\%           \\

\addlinespace
\bottomrule
\end{tabular}
\caption{Accuracy Comparison of HEC, Ensemble Methods, and the best individual base-learner (RoBERTa) on 20NewsGroups.}\label{pretrained_hec_vs_ensembles}
\end{table*}

\end{document}